\documentclass[conference]{IEEEtran}
\IEEEoverridecommandlockouts
\usepackage{cite}
\usepackage{amsmath,amssymb,amsfonts}
\usepackage{algorithmic}
\usepackage{graphicx}
\usepackage{textcomp}
\usepackage{xcolor}
\newcommand{\red}[1]{\textcolor{red}{#1}}
\newcommand{\green}[1]{\textcolor{green}{#1}}
\usepackage{booktabs}
\usepackage{pifont}
\newcommand{\cmark}{\green{\ding{51}}}%
\newcommand{\xmark}{\red{\ding{55}}}%
\usepackage{cleveref}
\usepackage{url}

\def\BibTeX{{\rm B\kern-.05em{\sc i\kern-.025em b}\kern-.08em
    T\kern-.1667em\lower.7ex\hbox{E}\kern-.125emX}}

\usepackage{fancyhdr,lipsum}
\fancypagestyle{firstpage}{
  \fancyhf{}
  \fancyhead[C]{To appear at the 2023 International Joint Conference on Neural Networks (IJCNN), Queensland, Australia, June 2023.}
  \fancyfoot[C]{\thepage}
}

\begin{document}

\title{SwiftTron: An Efficient Hardware Accelerator for Quantized Transformers}

\author{\IEEEauthorblockN{Alberto Marchisio\textsuperscript{1,*}\thanks{*These authors contributed equally to this work.}, Davide Dura\textsuperscript{2,*}, Maurizio Capra\textsuperscript{2,*}, Maurizio Martina\textsuperscript{2}, Guido Masera\textsuperscript{2}, Muhammad Shafique\textsuperscript{3}}
\IEEEauthorblockA{\textit{\textsuperscript{1}Technische Universit{\"a}t Wien, Vienna, Austria}\ \ \ \textit{\textsuperscript{2}Politecnico di Torino, Turin, Italy}\ \ \ \textit{\textsuperscript{3}New York University, Abu Dhabi, UAE}} 
\IEEEauthorblockA{\textit{Email: alberto.marchisio@tuwien.ac.at, s276493@studenti.polito.it, maurizio.capra@polito.it}}
\IEEEauthorblockA{\textit{maurizio.martina@polito.it, guido.masera@polito.it, muhammad.shafique@nyu.edu}}\\
\vspace*{-30pt}}


\maketitle
\thispagestyle{firstpage}

\begin{abstract}
Transformers' compute-intensive operations pose enormous challenges for their deployment in resource-constrained EdgeAI / tinyML devices. As an established neural network compression technique, quantization reduces the hardware computational and memory resources. In particular, fixed-point quantization is desirable to ease the computations using lightweight blocks, like adders and multipliers, of the underlying hardware. However, deploying fully-quantized Transformers on existing general-purpose hardware, generic AI accelerators, or specialized architectures for Transformers with floating-point units might be infeasible and/or inefficient.

Towards this, we propose \textit{SwiftTron}, an efficient specialized hardware accelerator designed for Quantized Transformers. \textit{SwiftTron} supports the execution of different types of Transformers' operations (like Attention, Softmax, GELU, and Layer Normalization) and accounts for diverse scaling factors to perform correct computations. We synthesize the complete \textit{SwiftTron} architecture in a $65$ nm CMOS technology with the ASIC design flow. Our Accelerator executes the RoBERTa-base model in $1.83\ ns$, while consuming $33.64\ mW$ power, and occupying an area of $273\ mm^2$. To ease the reproducibility, the RTL of our \textit{SwiftTron} architecture is released at \url{https://github.com/albertomarchisio/SwiftTron}.
\end{abstract}


\section{Introduction}

Among advanced Machine Learning (ML) models, Transformers are becoming mainstream for several applications like natural language processing and computer vision. However, they involve several compute-intensive operations 
like Multi-Head Self Attention and Layer Normalization with massive data streams. Hence, their execution is highly power-consuming when conducted on general-purpose hardware. Specialized architectures would be desirable to accelerate the execution of Transformers' operations and improve the energy-efficiency.

\begin{figure}[t]
\centering
\includegraphics[width=.9\linewidth]{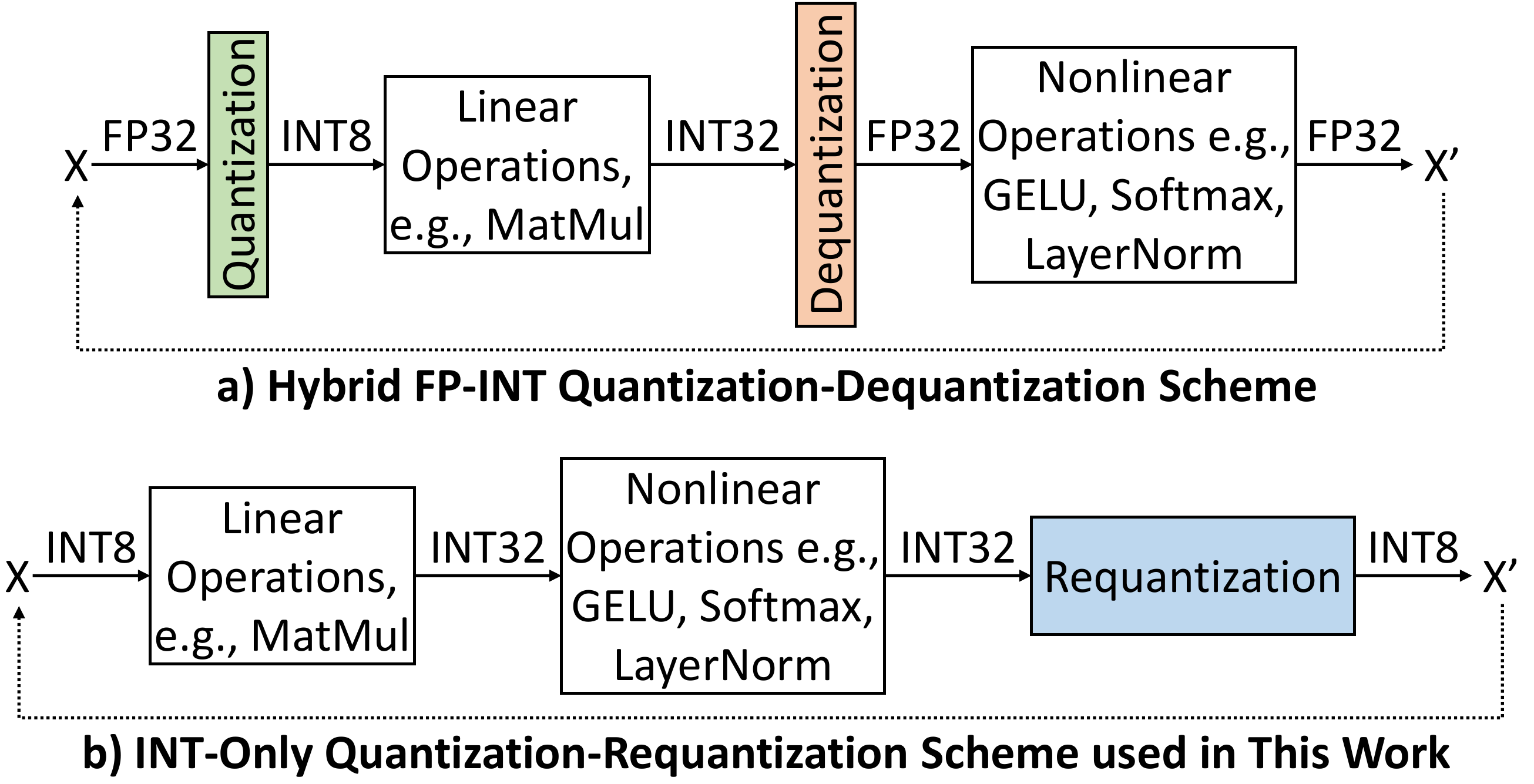}
\caption{Simplified diagrams showing the computation flows for \textbf{(a)}~the Quantization-Dequantization scheme of~\cite{lin2020towards}, and \textbf{(b)}~the quantization-based scheme adopted in this work that employs \textit{Requantization} blocks.}
\label{fig:Quantization_dequantization}
\vspace*{-10pt}
\end{figure}

Several types of ML accelerators have been recently integrated with the most common chips to execute massive matrix multiplications that are typical in convolutional and fully-connected layers~\cite{Shafique_2018_MLSurveyDATE}\cite{Marchisio_2019_DL4ECISVLSI}\cite{Capra_2020_UpdatedSurvey}\cite{Shafique_2021_ICCADSS}\cite{Dave_2022_VTSSS}. 
However, such generic architectures do not support some Transformer-specific operations, such as Attention, Softmax, Gaussian Error Linear Unit (GELU), and Layer Normalization. Hence, the neural accelerators need to be tailored for such unique operations involved in Transformers. Towards this, some architectures have recently been proposed. For instance, OPTIMUS~\cite{MLSYS2020_903ce922} optimizes the execution of matrix multiplications in transformers, and A\textsuperscript{3}~\cite{Ham_2020_A3} accelerates the execution of the Attention operation. However, these architectures only execute a certain part of a given Transformer, and do not provide a holistic acceleration platform for the complete Transformer. Note that other functions like Softmax, GELU, and Layer Normalization involve nonlinear operations that cannot be easily implemented in integer arithmetics. A common way of handling these operations is to iteratively quantize the input to compute matrix multiplication with integers and de-quantize the intermediate results to calculate the nonlinear operations in floating point arithmetic~\cite{lin2020towards}. A simplified scheme of this method is depicted in \Cref{fig:Quantization_dequantization}a. A similar approach is also adopted in the recent Transformer Engine of the NVIDIA H100 Tensor Core GPU~\cite{Nvidia_H100}, which implements the operations in FP8 arithmetic. However, this approach would require the design of several floating point arithmetic logic units with significant resource overheads compared to the integer counterparts.

\begin{figure}[t]
\centering
\includegraphics[width=.9\linewidth]{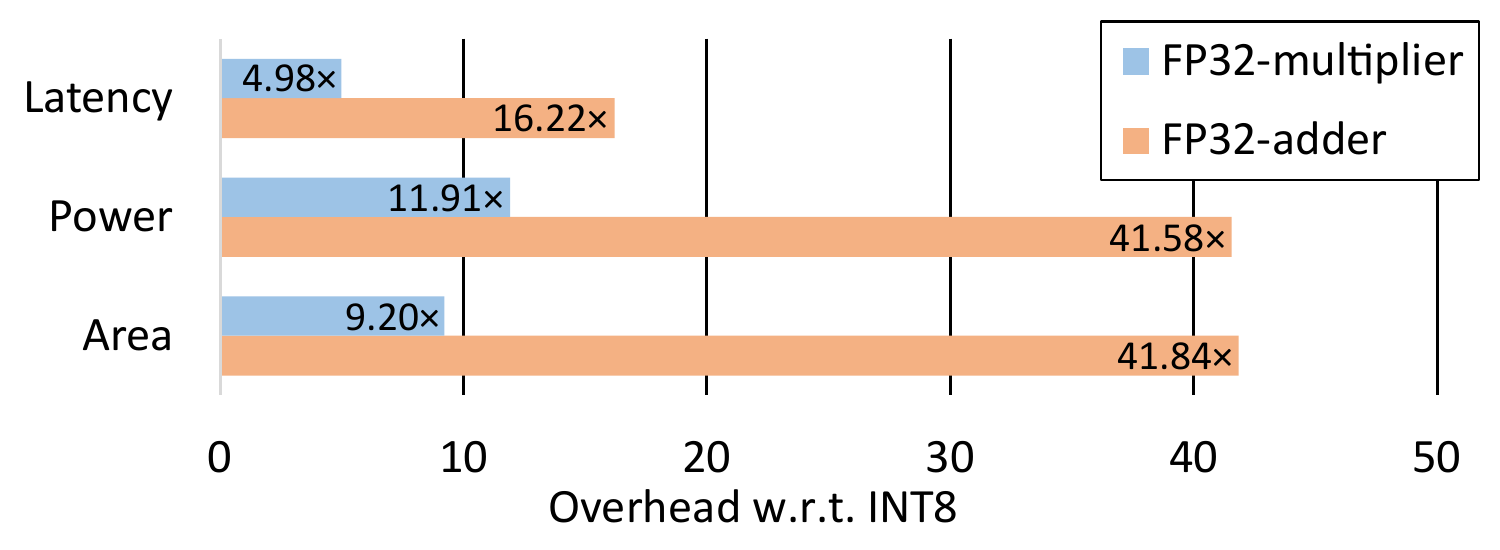}
\caption{Latency, power, and area overhead of a single adder and a single multiplier implemented in FP32 arithmetic, compared to their respective INT8 implementations.}
\label{fig:INT8_overhead}
\end{figure}

As a motivating experimental analysis comparing different arithmetics, we synthesize an INT8-adder, an INT8-multiplier, an FP32-adder, and an FP32-multiplier~\cite{Marcus_2004_FP32AdderMultiplier} in a 65 nm CMOS technology node with the Synopsys Design Compiler and evaluate latency, power, and area. \Cref{fig:INT8_overhead} shows the respective overhead of FP32 arithmetic operators compared to their respective INT8 implementations. These experiments show that the potential savings are about one order of magnitude.

On the other hand, computing nonlinear operations with quantized integers without incurring significant accuracy loss is nontrivial. The state-of-the-art in this regard is represented by the I-BERT~\cite{Kim2021iBert}, a framework that implements specific approximations to execute all the nonlinear operations of the Transformers with integer-only arithmetic. However, this work implemented the Transformers on general-purpose hardware, i.e., GPU. A specialized accelerator executing all the Transformer layers, including nonlinear operations, using only efficient integer arithmetic is missing, and highly required.



\subsection{Our Novel Contributions}

\begin{figure}[t]
\centering
\includegraphics[width=.99\linewidth]{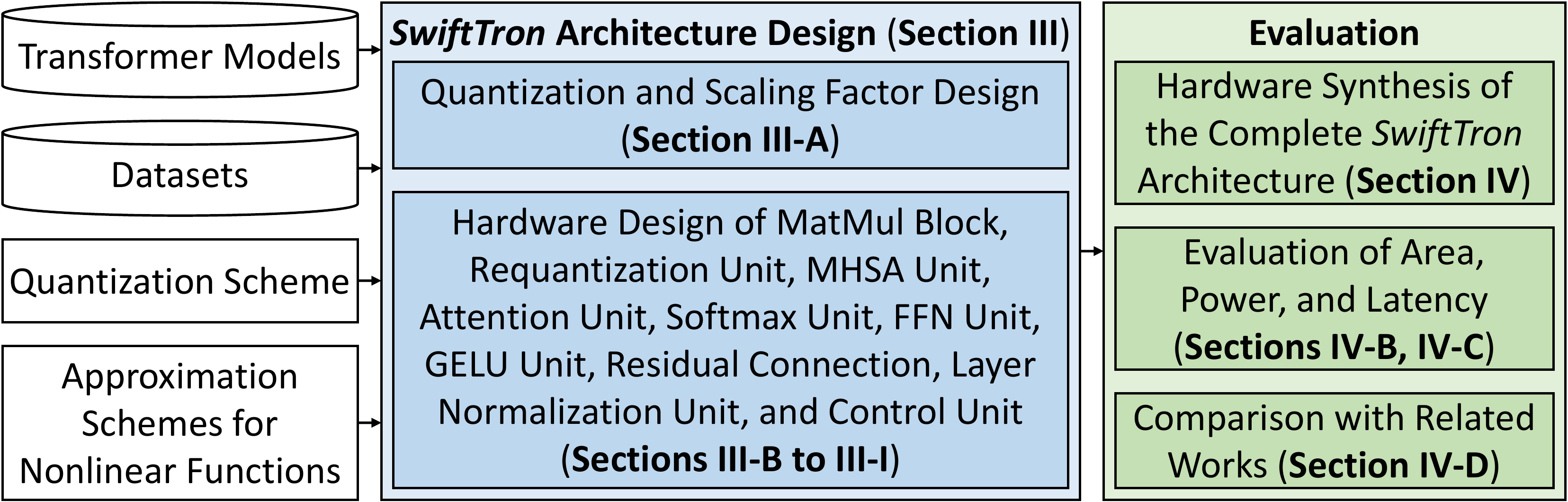}
\caption{Overview of our novel contributions in this work.}
\label{fig:Novel_contributions}
\vspace*{-10pt}
\end{figure}

The above-discussed limitations motivate us to propose \textit{SwiftTron}, an efficient hardware accelerator that executes quantized Transformers with integer-only operations (see \Cref{fig:Quantization_dequantization}b for our integer-only arithmetic flow), while focusing on multiple different operations of a given Transformer to provide a high degree of performance/energy efficiency. Our contributions are discussed in the following list (see \Cref{fig:Novel_contributions}).

\begin{itemize}
    \item We design the \textit{SwiftTron} hardware architecture, a specialized accelerator composed of several hardware units to execute different operations of the Transformers. (\textbf{\Cref{sec:SwiftTron_design}})
    \item We design and implement a quantization scheme for Transformers with scaling factors to correctly execute the linear operations with INT8 and the nonlinear operations with INT32 arithmetics. (\textbf{\Cref{subsec:Quantization_and_Scaling}})
    \item We synthesize the complete \textit{SwiftTron} architecture in a 65 nm CMOS technology node with the Synopsys Design Compiler and conduct gate-level simulations to measure the area and power consumption. (\textbf{\Cref{sec:Evaluation}})
    \item We compare the key features of our accelerator with the related work to highlight that our \textit{SwiftTron} is the first architecture that complies with all the desired features. (\textbf{\Cref{subsec:Comparison_Related_Work}})
    \item Towards encouraging fast advancements in the neural hardware accelerator and ML community, and to ease the reproducibility of our experiments, we open-source the complete RTL of our \textit{SwiftTron} accelerator architecture at \url{https://github.com/albertomarchisio/SwiftTron}.
\end{itemize}

\section{Background and Related Work}

\subsection{Transformer Models}
\label{subsec:background_transformers}

The Transformer network has been introduced in~\cite{vaswani2017attention}, which is the reference point of the related works and subsequent models. 
Transformers are formed of two main blocks, the encoder and the decoder. They are composed of the following layers\footnote{Note that, since there is little difference between the encoder and the decoder, the composition of their layers is very similar.}:

\begin{itemize}
    \item Multi-Head Self Attention (MHSA):
        \begin{itemize}
            \item Linear Transformation to compute Query ($Q$), Key ($K$), and Value ($V$) matrices for each head.
            \item Attention:
                \begin{itemize}
                    \item $Q \cdot K^T$ multiplication, where $K^T$ denotes the transposed matrix $K$
                    \item $Softmax(Q \cdot K^T) \cdot V$ multiplication
                \end{itemize}
            \item Linear Transformation after concatenating every head Attention output.
        \end{itemize}
    \item Residual Connection \& Layer Normalization in the MHSA
    \item Feed-Forward Network (FFN):
        \begin{itemize}
            \item Linear Transformation
            \item Activation Function
            \item Linear Transformation
        \end{itemize}
    \item Residual Connection \& Layer Normalization in the FFN
\end{itemize}

\begin{figure}[t]
\centering
\includegraphics[width=\linewidth]{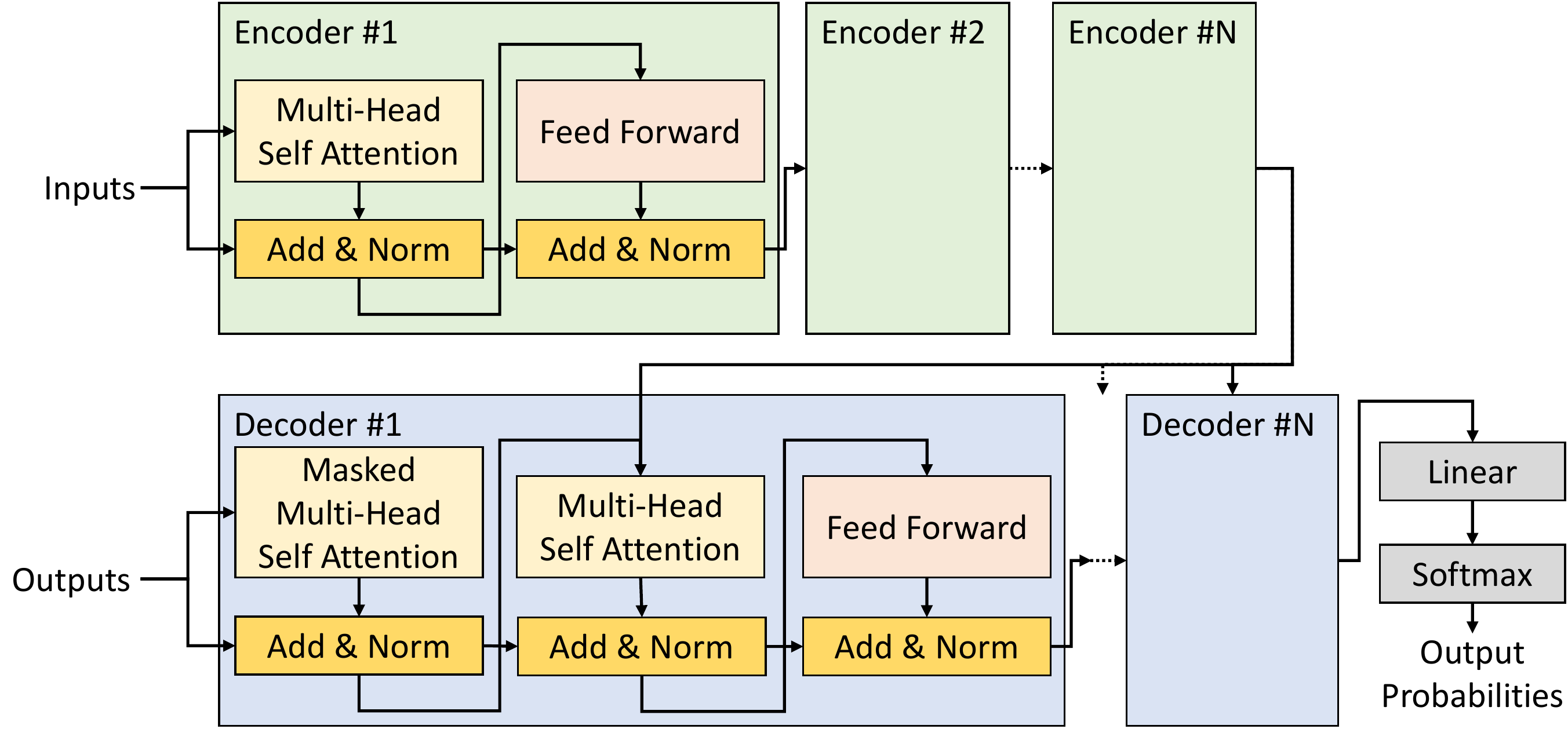}
\caption{Architectural model of the Transformer~\cite{vaswani2017attention}, where inputs and outputs are taken after the positional encoding operations.}
\label{fig:Transformer_model}
\vspace*{-10pt}
\end{figure}

When repeated multiple times, these layers form the structure of the encoder and the decoder, as shown in \Cref{fig:Transformer_model}. Note that the typical activation functions for Transformers\footnote{There exist a few other options, but these are the most commonly used activation functions in Transformers.} are the \textit{ReLU}~\cite{Nair2010ReLU} and the \textit{GELU}~\cite{Hendrycks2016GELU}.
The baseline Transformer~\cite{vaswani2017attention} is composed of $N=6$ Encoder layers and $N=6$ Decoder layers, while many different architectures have been proposed. The work of~\cite{dehghani2018universal} proposed a tunable Transformer model where the same Encoder/Decoder layer is used until the computation reaches the desired results. This design led to having only one instance of these layers and the same parameter set for each iteration, which either can be fixed a priori or adapted dynamically during inference.
On the other hand, the BERT-like architectures~\cite{devlin2018bert} rely only on the Encoder part of the network, achieving state-of-the-art performances in several tasks.

The \textit{Attention} is the key operation of a Transformer, unleashing overall better performance than other ML architectures. However, due to the necessary non-linearities, its compute-intensive operations challenge typical neural hardware accelerators. Moreover, Transformers' massive sizes, like the OpenAI GPT-3~\cite{Brown2020GPT3} with $96$ hidden layers and $175\ B$ parameters, lead to large memory footprint, latency, and power consumption, making their hardware execution extremely resource-demanding.

Some works~\cite{Wang2020Linformer}\cite{Liu2021SwinTransformer} proposed mathematical manipulations for reducing the complexity of the needed operators in functions like Softmax and Layer Normalization. Other works inspired by compression techniques like pruning or knowledge distillation reduce the model's operations and parameters~\cite{Fedus2021SwitchTransformer}. \textit{Note that these approaches are orthogonal to our work.}

A potential technique that has the higher impact on implementing Transformer networks onto hardware devices is \textit{quantization}, which transforms the floating-point values, universally used in the Transformer models, into integer values while trying to minimize the consequent precision loss. For instance, the I-BERT~\cite{Kim2021iBert} implements the entire BERT network~\cite{devlin2018bert} with integer operations. This process consists of an efficient way to have simpler operators and lighter number representation, helping both the resources and memory constraints for an efficient accelerator design. Similarly, the I-ViT~\cite{Li2022iViT} quantizes the Vision Transformer~\cite{Dosovitskiy2021ViT} for image classifiation. \textit{However, these works are deployed on GPUs, while our focus is on specialized hardware accelerators.}

\subsection{AI Hardware Accelerators Executing Transformers' Operations}

Generic AI hardware accelerators are based on arrays of Processing Elements (PEs)~\cite{Capra2020AcceleratingDNNSurvey}. Adapting an existing accelerator to compute the dot product between the query, key, and value matrices in the MHSA mechanism would require several modifications in its architecture and would not be straightforward. Therefore, designing specialized hardware architectures for Transformers is highly desirable.

Recent works proposed hardware designs for executing some of the Transformer's operations efficiently. However, they primarily use floating-point representation, which is complex and expensive in terms of hardware resources.
To the best of our knowledge, there are no accelerators in prior works implementing the entire Transformer network using only integer computations and simple approximations for nonlinear operations. Still, the work in~\cite{lu2020hardware} presents an interesting hardware architecture for MHSA and FFN. Besides several novel additions, our work also takes some inspiration from these works regarding the column-oriented computations. In fact, \textit{our data flow is designed to use one column at a time of the matrices under processing. Since the matrix multiplication, which is the most used operation in Transformers, requires the input to be read column-by-column according to its algorithm, this data flow structure enables a simple interface between blocks.}
The architecture proposed in~\cite{MLSYS2020_903ce922} proposes an optimization for the hardware matrix multiplication in the Transformer. The work in~\cite{wang2020hat} analyzes the network design considering the hardware latency in the process. The work in~\cite{wang2020efficient} evaluates the performance of a floating-point implementation on CPUs and GPUs.


The main drawbacks of these related works from the hardware execution perspective are the following:

\begin{enumerate}
    \item The Floating Point representation is employed in some computations~\cite{lin2020towards}\cite{bhandare2019efficient}\cite{https://doi.org/10.48550/arxiv.1910.10485}. This is a simulated quantization process where almost every variable is quantized. However, the GELU, Softmax, and LayerNorm are computed with floating-point operations. It not only increases the hardware complexity, but it also requires both quantization and dequantization layers to convert data between blocks. 
    \item Complex operators like exponential and square root are implemented with expensive LUTs~\cite{Xin_2022_EFATrans}\cite{Liu_2021_FQBERT}, or using different approaches that involve FFTs~\cite{Li_2020_FTRANS}.
\end{enumerate}

\textit{To overcome these issues, our proposed \textit{SwiftTron} architecture executes all the layers and functions in transformers using only integer computations.}

\section{SwiftTron: Hardware Architecture Design} 
\label{sec:SwiftTron_design}

\begin{figure}[t]
\centering
\includegraphics[width=.85\linewidth]{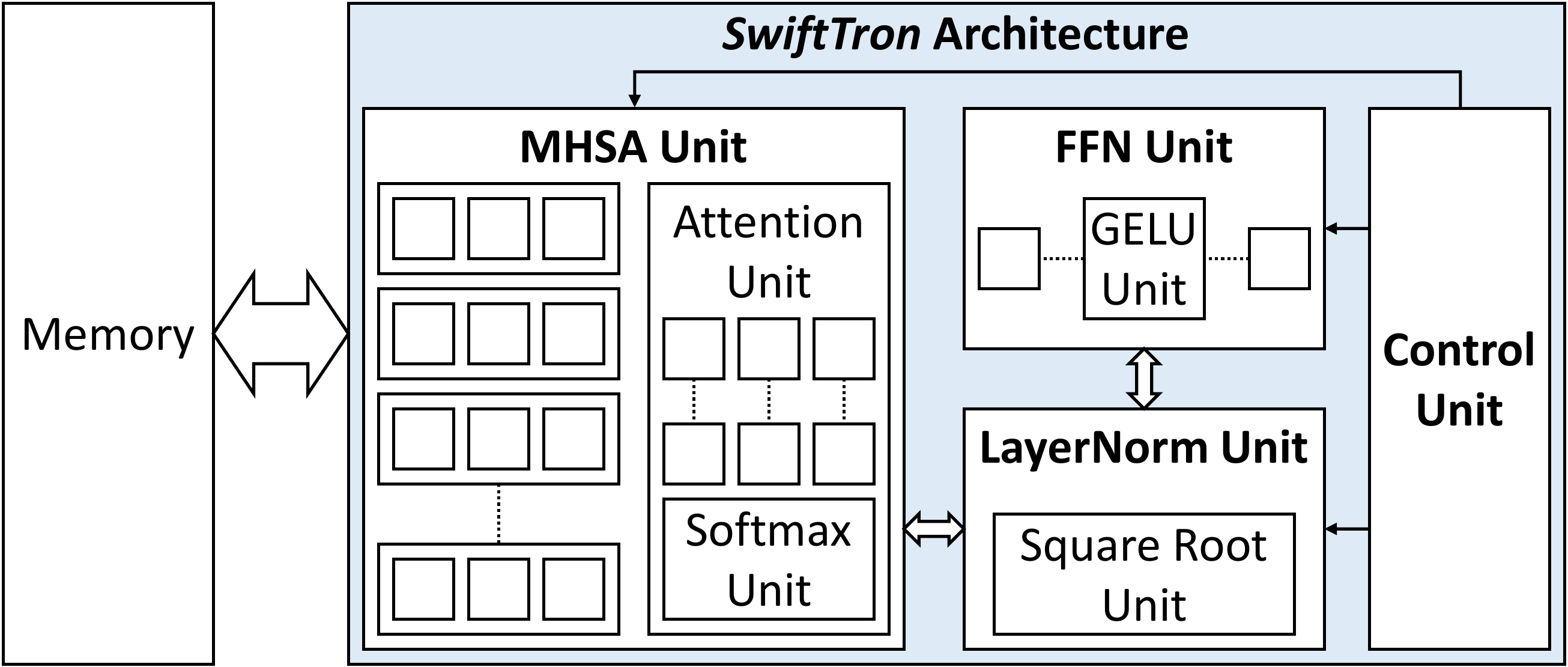}
\caption{Top-level overview of our \textit{SwiftTron} architecture.}
\label{fig:Top_level_architecture}
\vspace*{-10pt}
\end{figure}

This section describes each hardware block designed for executing the Transformers' layers (as discussed in \Cref{subsec:background_transformers}). The layers are mainly composed of linear transformations and nonlinear functions.
Since the linear layers are based on matrix multiplications, a \textit{MatMul} block is designed. 
On the other hand, to execute nonlinear functions, namely the \textit{Softmax}, \textit{GELU}, and \textit{Square Root} for the Normalization, our work deploys second-order polynomial approximations and recursive implementation, which are based on the concepts from~\cite{Kim2021iBert}. A top-level view of our proposed \textit{SwiftTron} architecture is shown in \Cref{fig:Top_level_architecture}.

The Transformer's main parameters are the model dimension $d$, the number of heads $k$, the sentence length $m$, and the feed-forward dimension $d_{ff}$ (usually two or four times the value of $d$).

\subsection{Quantization and Scaling Factor Design}
\label{subsec:Quantization_and_Scaling}

Before going into detail of each component, it is important to highlight that quantized values have their corresponding scaling factors derived during the quantization process. 

Formally, given $a$ a floating-point value and $q_a$ its quantized value, the scaling factor $S_a$ is defined such that $a = Q_a S_a$. Scaling factors allow the transformations from floating point to integer and vice versa, and determine the correct operation between two integers. For example, two numbers with different scaling factors cannot be directly summed together, but an extra component is needed for the \textit{Residual Connection}. Furthermore, the scaling factors are fundamental for computing the algorithm's coefficients, especially in the nonlinear functions. 
Some constraints on the scaling factor values are applied to have representable coefficients and to limit the risk of overflow. A scaling factor is a floating-point number that is not directly included in the architecture to avoid FP operators, but its value is fixed for each layer at design time.
Another critical aspect to consider is the data representation in the architecture. The matrix multiplications are conducted with INT8 inputs and INT32 accumulators. The nonlinear functions operate on INT32 to avoid excessive accuracy loss. Therefore, a \textit{Requantization} block is needed to bring the INT32 values back to INT8 as input to the subsequent MatMul operations.

Dealing with the scaling factor in linear operations is straightforward. For instance, the multiplication between to numbers ($a$ and $b$) with different scaling factors is defined as $a \cdot b = q_a S_a \cdot q_b S_b = (q_a \cdot q_b) (S_a \cdot S_b)$. This property holds for matrix multiplications (MatMul), since its resulting expression is $MatMul (Sq) = S \cdot MatMul(q)$.

However, transformers contain several nonlinear operations, such as GELU, Softmax, and Layer Normalization, for which this property does not hold. For this reason, these operations can be either approximated using a second-order polynomial, as in the case of GELU and Softmax, or computed iteratively, like for the square root of the Layer Normalization.

\subsection{Hardware Architecture of the MatMul Block}

\begin{figure}[t]
\centering
\includegraphics[width=.8\linewidth]{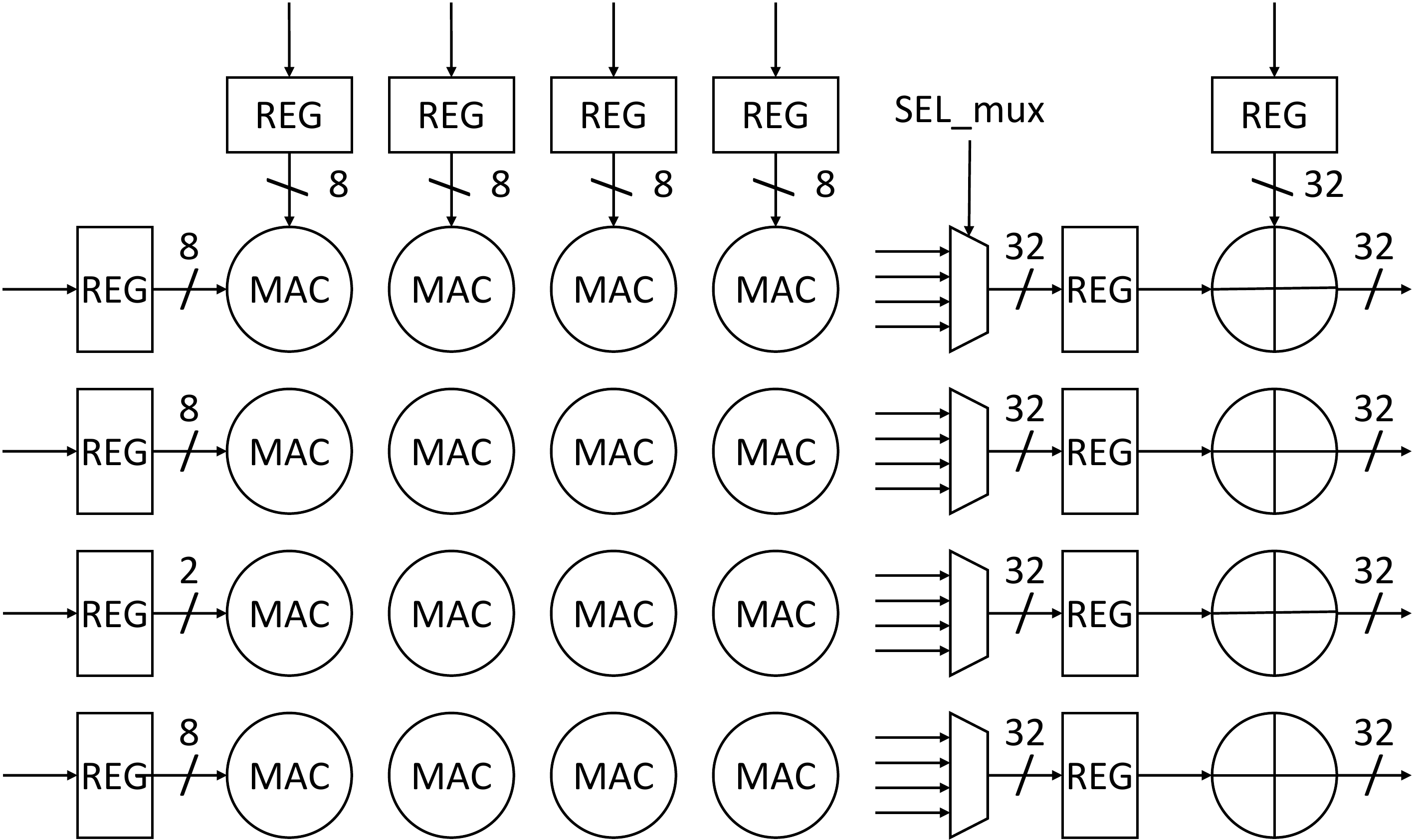}
\caption{Architectural view of a $4\times4$ MatMul block with bias.}
\label{fig:MatMul}
\end{figure}

The MatMul block is extremely important since it is used extensively in the following designs. It is formed by $\#rows$~$\times$~$\#columns$ Multiply-and-Accumulate (\textit{MAC}) elements that, at every iteration, receive their corresponding row and column input and update the accumulator. \Cref{fig:MatMul} reports an example with a $4 \times 4$ MAC array. After all the inputs are scanned, they store the final output that can be read column-by-column by tuning the selector of the output multiplexer.
In the following layers, several MatMul operations are required with different dimensions, but depending on what type of Transformer workload to execute, these components can be shared and/or reused.

The bias addition for the linear transformations can be incorporated into the component, like in this example. The bias is added when reading the output matrix. Hence, a different value is added to each column. For multiplications that do not require bias, this process can be ignored.

\subsection{Hardware Architecture of the Requantization Unit}

A scaling factor change is necessary to reduce the precision from $32$ bits to $8$ bits. To perform this transformation, the \textit{Dyadic Numbers} concept~\cite{yao2021hawq} is involved.
Starting from the $32$-bit representation of a value $a$, denoted with its quantized value $q_a$ and its quantization scale $S_a$ such that $a = q_aS_a$, the final representation should be $o = q_oS_o$ with $q_o$ on $8$ bits. Hence, equalling $a$ and $o$ since the real value must remain unchanged, the formula is derived in \Cref{eq:req_example}.

\begin{equation}
    \label{eq:req_example}
    q_aS_a = q_oS_o \longrightarrow q_o = q_a\frac{S_a}{S_o}
\end{equation}

Remembering that the scaling factors are not strictly integers but can assume any real value, this expression cannot be implemented directly on integer-only resources.
Hence, the scaling factor ratio is represented with a dyadic number, a rational number in the format of $b/2^c$, where $b$ and $c$ are two integer numbers.
The final expression is shown in \Cref{eq:req_dyadic}.

\begin{equation}
    \label{eq:req_dyadic}
    q_o = q_a\frac{S_a}{S_o} = q_a DN(\frac{S_a}{S_o}) = q_a * \frac{b}{2^c}
\end{equation}

\begin{figure}[t]
\centering
\includegraphics[width=.4\linewidth]{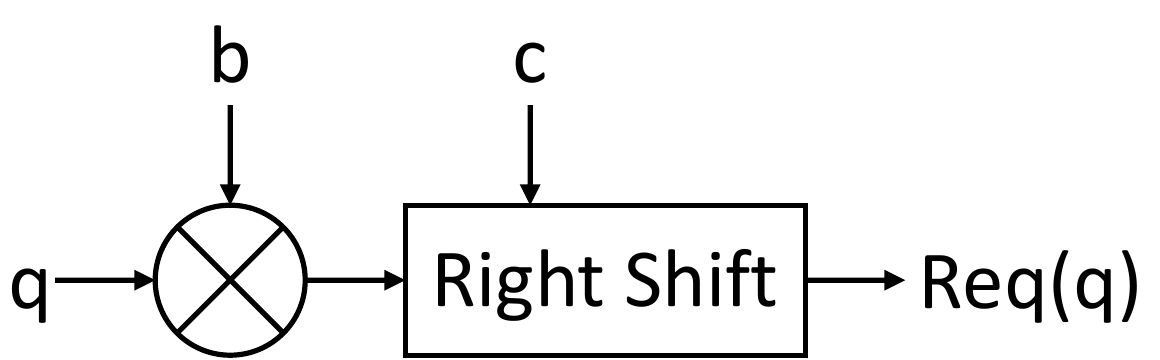}
\caption{Architecture details of the Requantization (Req) unit.}
\label{fig:Requantization}
\vspace*{-10pt}
\end{figure}

This convention also avoids the need to use dividers, as the required resources are only an INT32 multiplication and a one-bit shifting, as shown in \Cref{fig:Requantization}.

\subsection{Hardware Architecture of the MHSA Unit}

\begin{figure}[t]
\centering
\includegraphics[width=.95\linewidth]{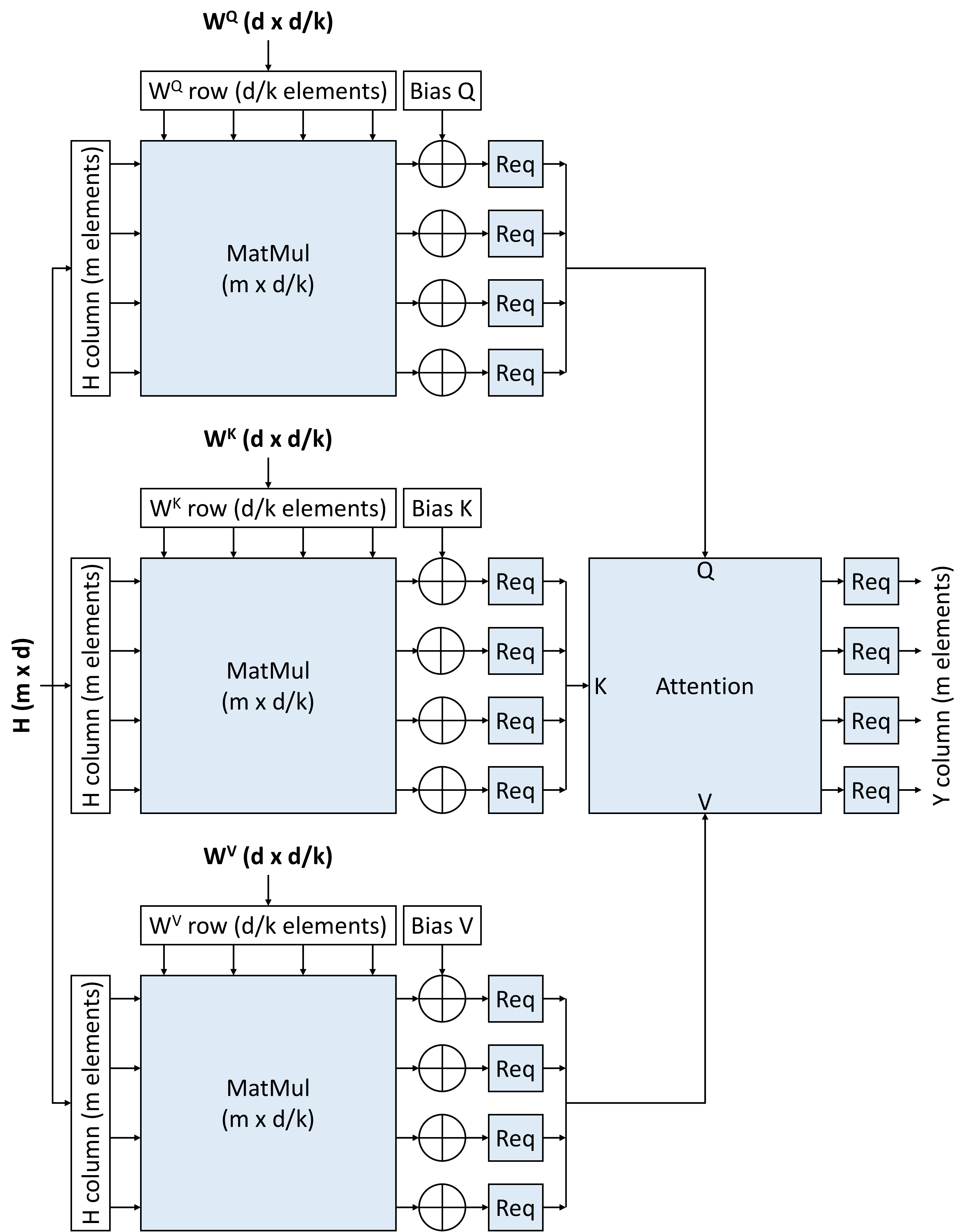}
\caption{Architecture details of a single head in the MHSA, composed of three MatMul blocks, performing computations on Query ($Q$), Key ($K$), and Value ($V$) matrices, and one Attention block.}
\label{fig:MHSA_one_head}
\end{figure}

\begin{figure}[t]
\centering
\includegraphics[width=.85\linewidth]{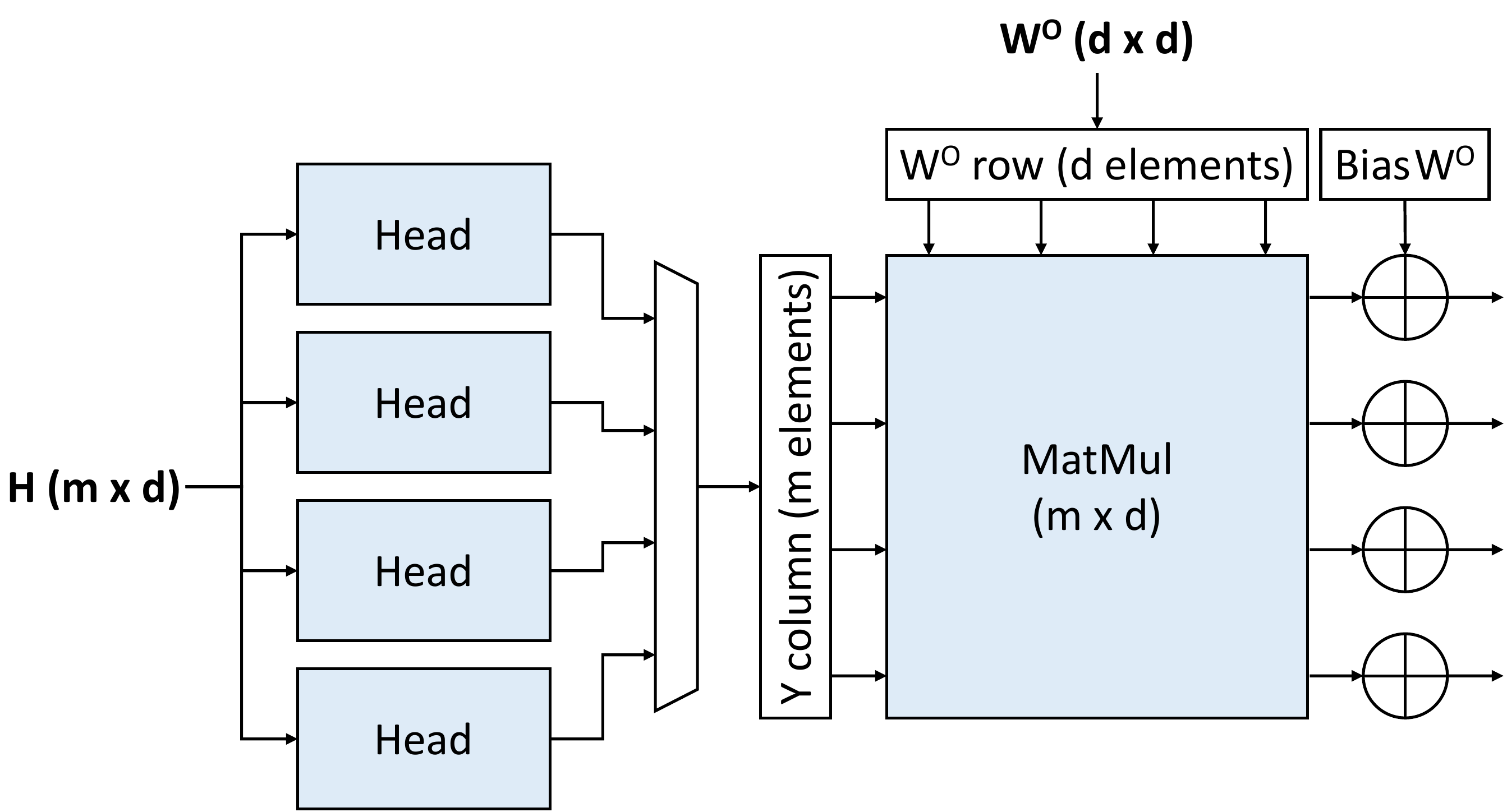}
\caption{Architecture details of the complete MHSA unit with multiple heads. The example in the figure contains $4$ heads and another MatMul block that generates the outputs.}
\label{fig:MHSA_multi_head}
\vspace*{-10pt}
\end{figure}

The MHSA block is responsible for computing the correspondent operation in a Transformer. A single head, whose architecture is shown in \Cref{fig:MHSA_one_head}, is composed of three MatMul blocks with inputs connected to Query, Keys, and Values, and an Attention operator. \Cref{fig:MHSA_multi_head} shows an example of the complete MHSA architecture composed of $4$ heads and another MatMul block that generates the outputs.

The choice of the number of heads to be computed in parallel depends on the available hardware resources. Different architectural configurations can be designed, from processing one head at a time to computing all heads concurrently. Therefore, data can be either processed concurrently or sequentially by reusing the head blocks. Consequently, the computations of the final MatMul can be conducted with multiple batches of data. Whenever it receives as input one head output (coming in order), it updates its accumulators.

\subsection{Hardware Architecture of the Attention Unit}

\begin{figure}[t]
\centering
\includegraphics[width=.99\linewidth]{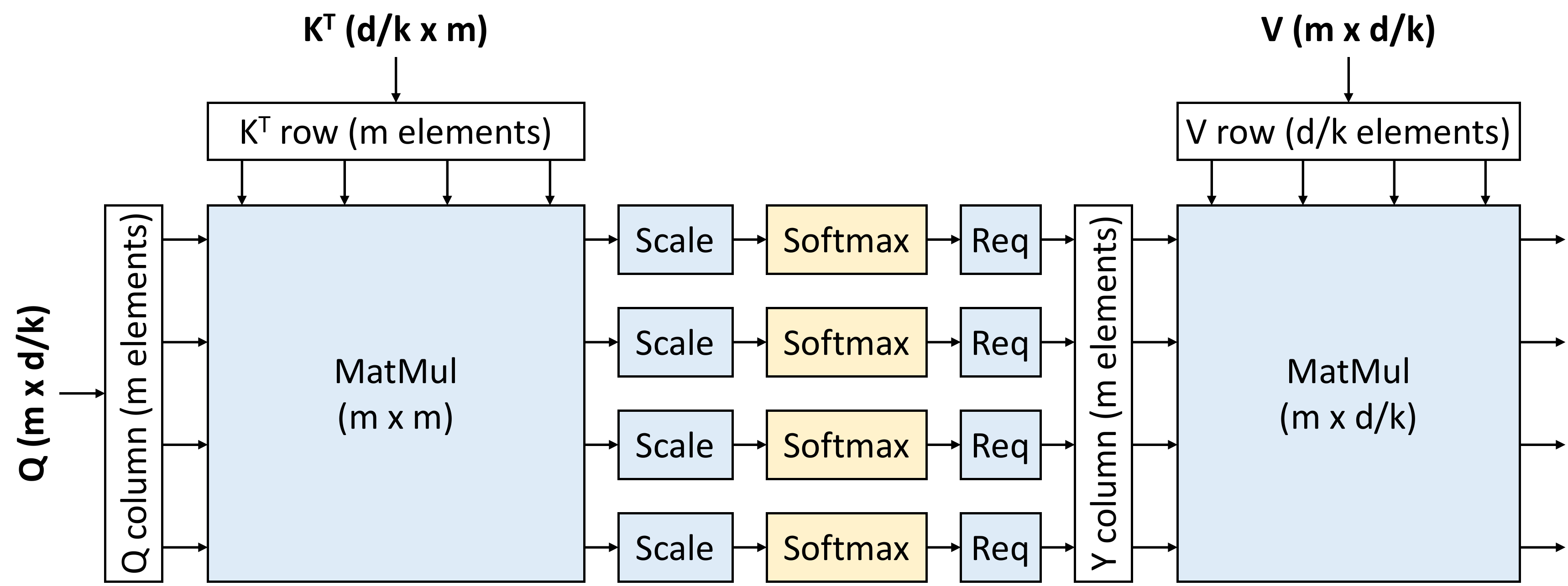}
\caption{Attention architecture, composed of MatMul, Scale, Softmax, and Requantization operators.}
\label{fig:Attention}
\end{figure}

The Attention architecture is shown in \Cref{fig:Attention}. It is composed of two plain MatMuls with a \textit{Softmax} in between. The \textit{Scale} simply consists of a division by the model dimension $d$. If the value of $d$ is a multiple of $2$, it becomes a simple shift operation. The \textit{Requantization} is required to keep the input of the second MatMul to INT8.

\subsection{Hardware Architecture of the Softmax Unit}

\begin{figure}[t]
\centering
\includegraphics[width=.99\linewidth]{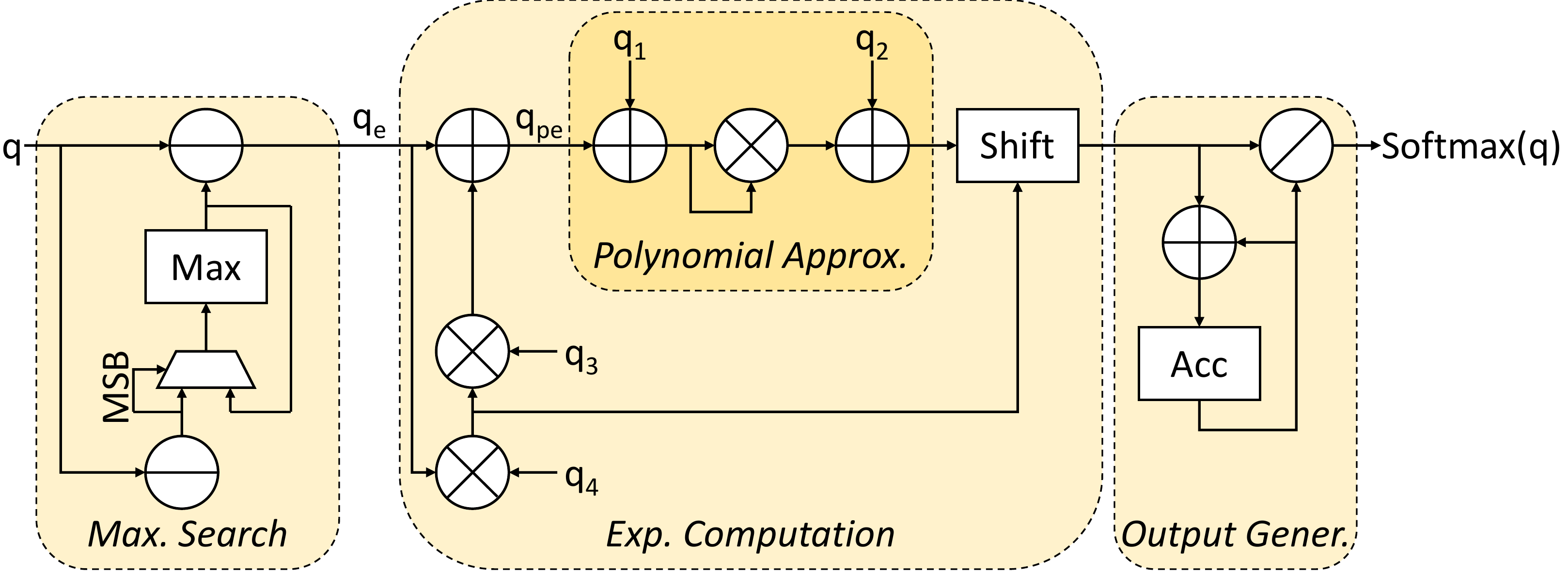}
\caption{Architecture of the Softmax operator. According to~\cite{Kim2021iBert}, \mbox{$q_1 = \lfloor b / S_{pe} \rfloor$}, \mbox{$q_2 = \lfloor c / a S_{pe}^2 \rfloor$}, \mbox{$q_3 = \lfloor ln2 / S_e \rfloor$}, and \mbox{$q_4 = \lfloor - 1 / q_3 \rfloor$}, where $S_e$ is the scaling factor of the exponential computation input (relative to $q_e$), $S_{pe}$ is the scaling factor of the polynomial approximation input (relative to $q_{pe}$) and $a, b, c$ are the coefficient of the second-order polynomial that approximates the function, i.e., \mbox{$a(x+b)^2+c$}. Therefore, $q_{1,2,3,4}$ can be computed at design time and provided as constant values to the \textit{SwiftTron} architecture.}
\label{fig:Softmax}
\vspace*{-10pt}
\end{figure}

Since the Softmax is performed along the rows of the $Q \times K^T$ matrix, $m$ Softmax components are instantiated and work concurrently. A single Softmax operator is shown in \Cref{fig:Softmax}. Its computation requires three phases, namely maximum search, exponential computation, and output generation. 

\begin{figure}[t]
\centering
\includegraphics[width=.99\linewidth]{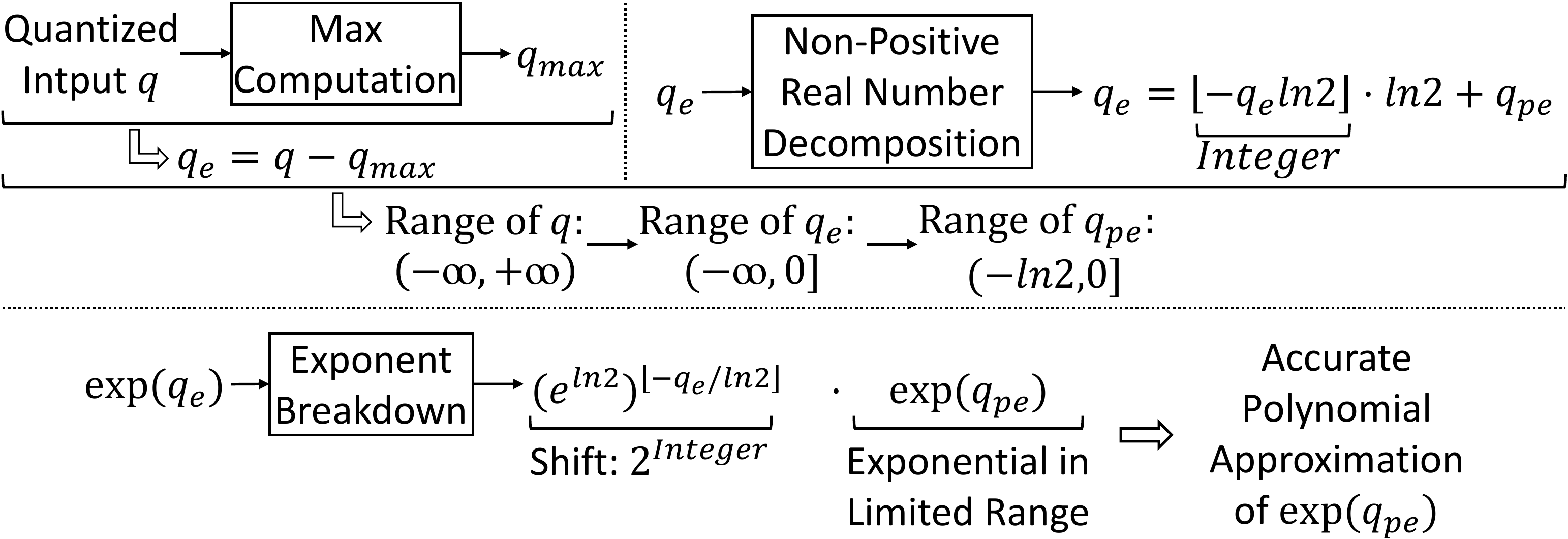}
\caption{Approach used for manipulating the Softmax operator to allow a second-order polynomial approximation of the exponential function in a restricted range of values.}
\label{fig:Softmax_algorithm}
\end{figure}

The implementation approach of the Softmax unit (see \Cref{fig:Softmax_algorithm}) aims at restricting the range of values in which the exponential function needs to be computed. Following the property described in \Cref{eq:softmax_max}, as in the Softmax inputs are limited by their maximum value, subtracting the maximum value leads to dealing with non-positive real numbers, which can be decomposed~\cite{Kim2021iBert}. Consequently, the exponential function must be computed only for the restricted range of \mbox{$[- ln2,\ 0]$} and can be approximated with a second-order polynomial.

\vspace*{-10pt}

\begin{equation}
    \resizebox{.91\columnwidth}{!}{%
    $Softmax(\textbf{x}_i) = \frac{exp(x_i)}{\sum_{j=1}^k exp(x_j)} = \frac{exp(x_i-x_{max})}{\sum_{j=1}^k exp(x_j-x_{max})}$
    }
    \label{eq:softmax_max}
\end{equation}

It is evident that with this approximation, only simple operators are involved, like a comparator for the maximum, adders, and multipliers. The most complex operator is the divider, whose implementation consumes relatively more resources.

\subsection{Hardware Architecture of the Feed-Forward Network Unit}

\begin{figure}[t]
\centering
\includegraphics[width=.99\linewidth]{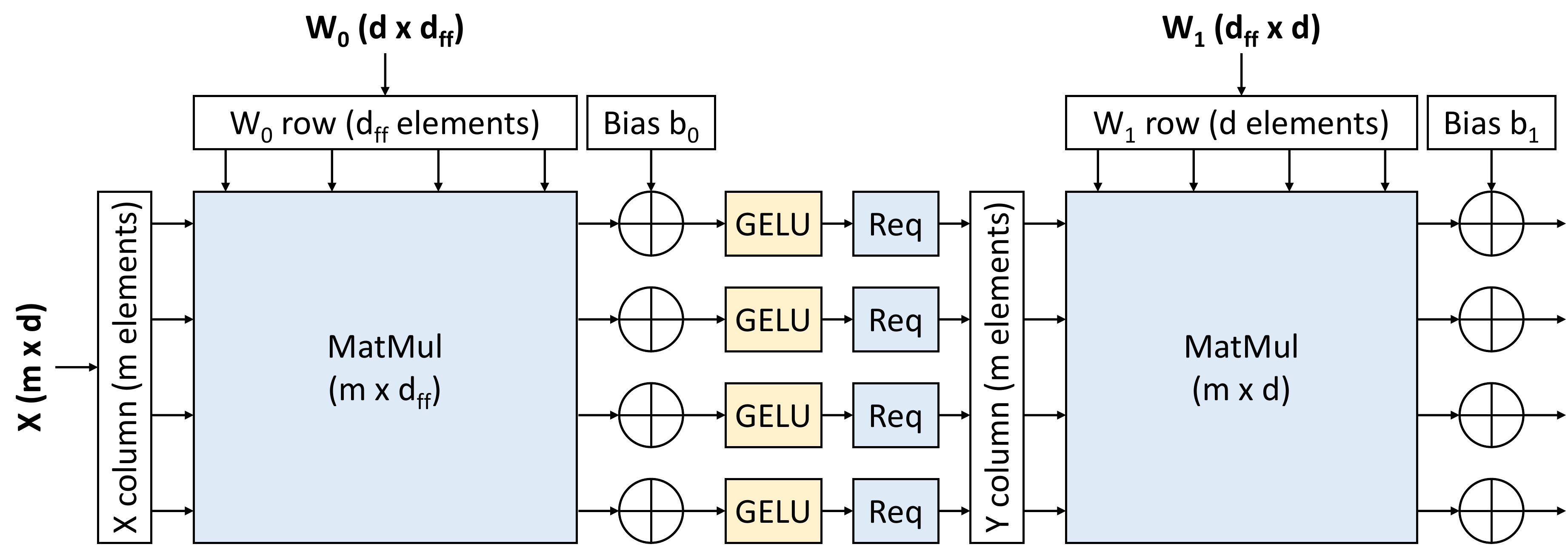}
\caption{Architecture of the Feed-Forward Network unit, composed of MatMul, GELU, and Requantization operators.}
\label{fig:FeedForwardNetwork}
\vspace*{-10pt}
\end{figure}

This sub-layer has two linear transformations separated by an activation function. The transformations, implemented with MatMul blocks, are the biggest of the Transformer architecture, as $d_{ff}$ is usually $4 \times$ the dimension $d$. The FFN architecture is depicted in \Cref{fig:FeedForwardNetwork}. The activation function used in our architecture is the GELU~\cite{Hendrycks2016GELU}, which despite being more complex, has better performances than the ReLU.

\subsection{Hardware Architecture of the GELU Unit}

\begin{figure}[t]
\centering
\includegraphics[width=.85\linewidth]{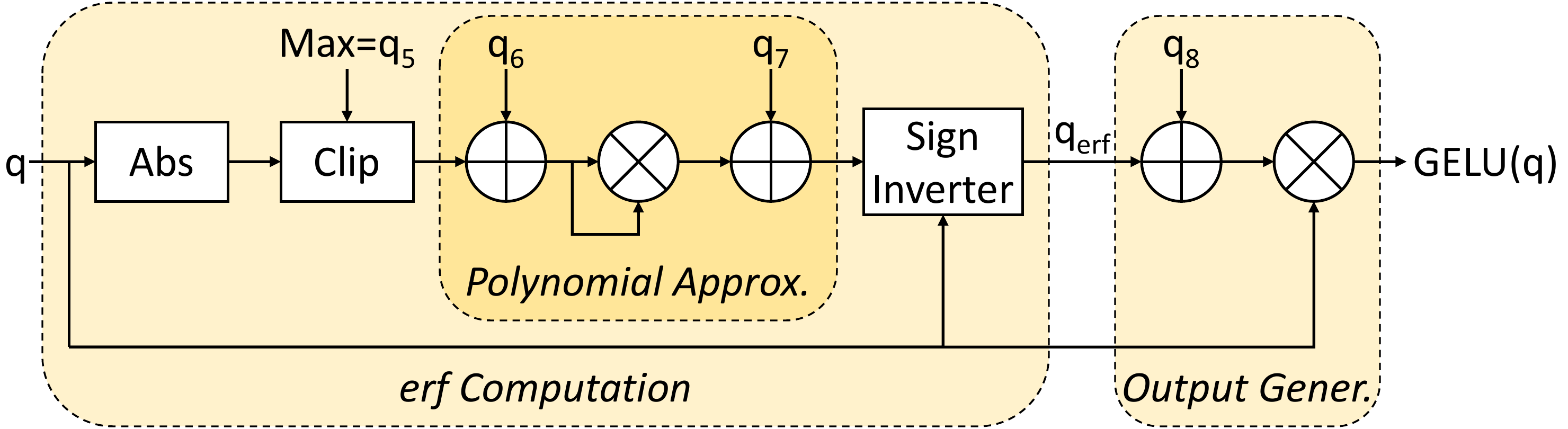}
\caption{Architecture of the GELU operator. According to~\cite{Kim2021iBert}, \mbox{$q_5 = - b / S$}, \mbox{$q_6 = \lfloor b / S \rfloor$} \mbox{$q_7 = \lfloor c / a S^2 \rfloor$}, and \mbox{$q_8 = \lfloor 1 / S_{erf} \rfloor$}, where $S$ is the scaling factor of the GELU input (relative to $q$), $S_{erf}$ is the scaling factor of the error function output (relative to $q_{erf}$) and $a, b, c$ are the coefficient of the second-order polynomial that approximates the function, i.e., \mbox{$a(x+b)^2+c$}. Therefore, $q_{5,6,7,8}$ can be computed at design time and provided as constant values to the \textit{SwiftTron} architecture.}
\label{fig:GELU}
\end{figure}

The GELU operator, shown in \Cref{fig:GELU}, contains the computation of the error function (\textit{erf}). This nonlinear function is linearized through another second-order polynomial with limited input. With this approximation, the resulting operators are only adders and multipliers, with some sign-handling operations that complete the execution.

\subsection{Hardware Architecture of the Residual Connection and Layer Normalization Units}

Since the MHSA and FFN are residual blocks, their outputs are added to the original inputs. As we are dealing with quantized values, the two addends need to have the same scaling factor before being added together. This transformation is achieved using a Dyadic unit, already discussed in the Requantization unit (recall \Cref{eq:req_dyadic}), implemented with a multiplication by a coefficient and a right-shifting. It is a combinatorial block that is replicated by the number of rows, as it receives one column at a time coming from the previous sub-blocks.

\begin{figure}[t]
\centering
\includegraphics[width=.99\linewidth]{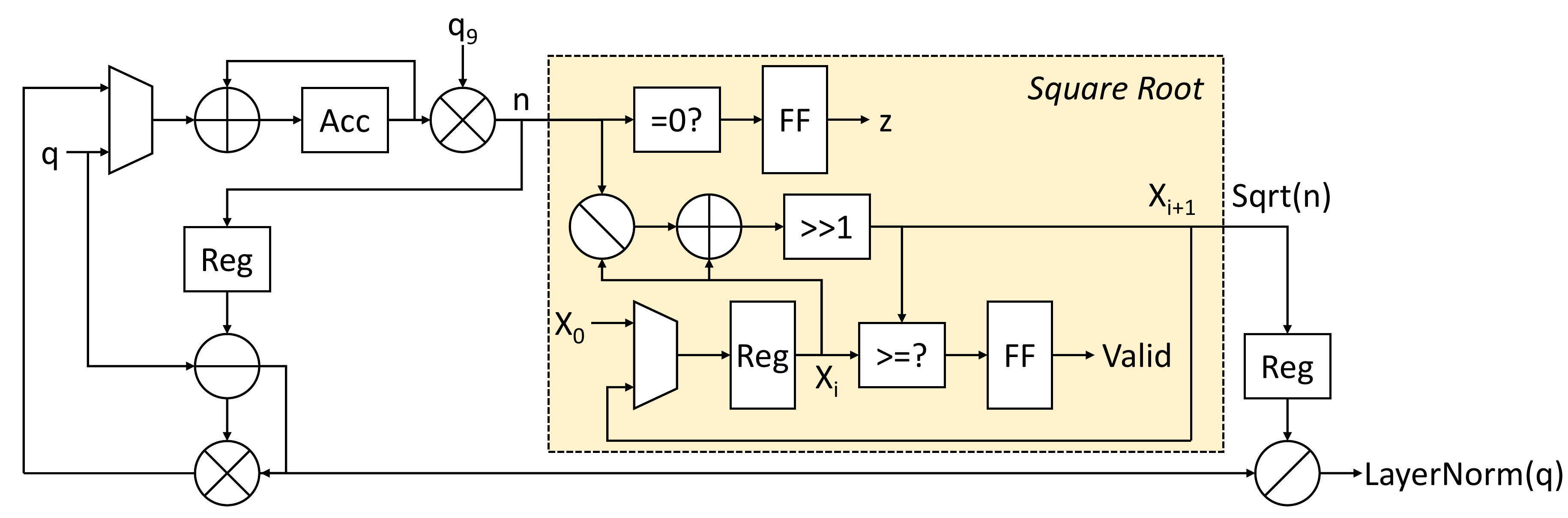}
\caption{Layer Normalization architecture, where the Square Root unit implements a recursive algorithm. The $Valid$ and $z$ signals are flags for assisting the control unit with the correct timing.}
\label{fig:LayerNorm}
\end{figure}

After the residual connection, the Layer Normalization (LayerNorm) is required. Its architecture is depicted in \Cref{fig:LayerNorm}. Similarly to the Softmax component, since the LayerNorm operation works on the row elements, $d$ instantiations are needed. Moreover, it is composed of three phases, the mean value calculation, the standard deviation calculation, and the output generation. 

The only nonlinear operation is the square root, which is implemented using an iterative algorithm as proposed in~\cite{Crandall_2006_PrimeNubmers}, also adopted in~\cite{Kim2021iBert}. It is a recursive algorithm that needs multiple cycles to compute the output. It includes combinatorial operators and registers to store intermediate values and break the loop. 

It has a constant initial value, defined as $x_0$. At every iteration, the partial result $x_i$ is compared to the partial result of the next iteration $x_{i+1}$ that is equal to $(x_i + x_i/n)/2$. Note that the division by $2$ is implemented through a simple one-position right shift. The algorithm iterates until $x_{i+1}$ is larger than or equal to $x_i$. When this happens, the final result is saved into the dedicated register. Since the number of cycles needed is unknown a priori, the $Valid$ and $z$ signals are flags for assisting the control unit and generate the correct timing. In the special case when the square root input is zero, the output goes directly to zero, and no iterations are needed.

\subsection{Hardware Architecture of the Control Unit}

\begin{figure}[t]
\centering
\includegraphics[width=.85\linewidth]{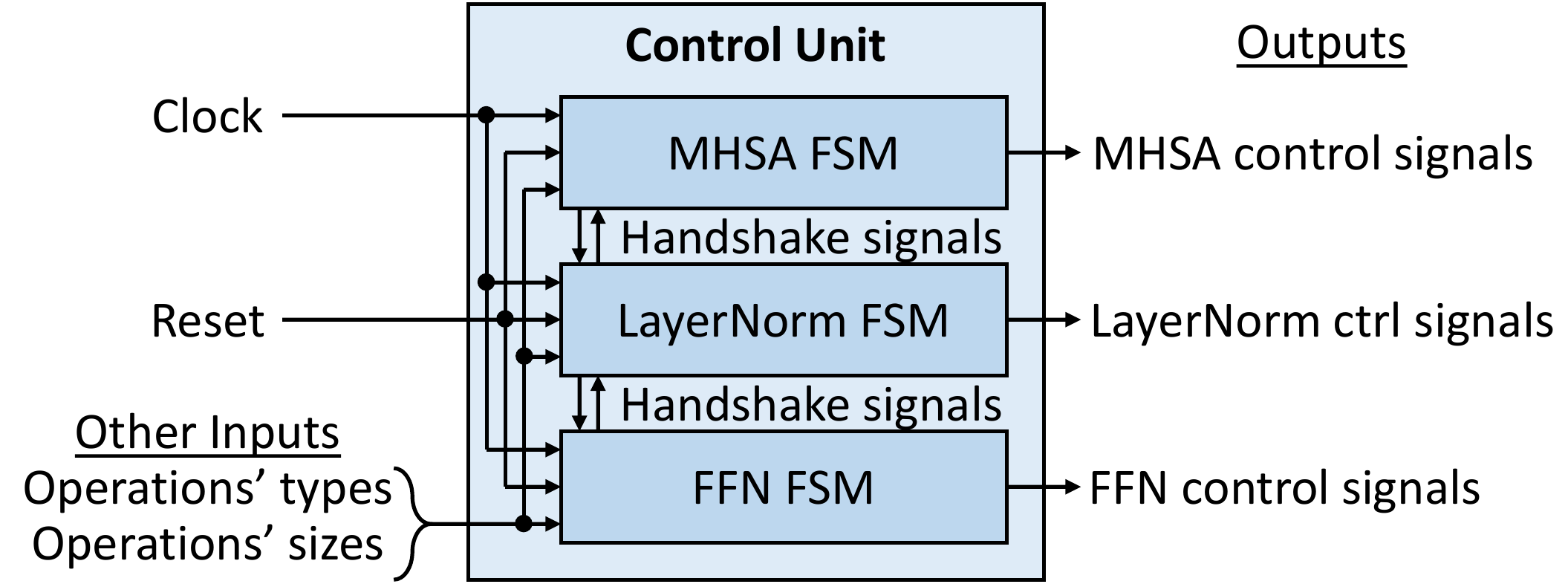}
\caption{Functionality of the control unit in our \textit{SwiftTron} architecture, composed of dedicated Finite State Machines for MHSA, LayerNorm, and FFN operations.}
\label{fig:Control_Unit}
\end{figure}

At each stage of the Transformers' process, the control unit generates different control signals for all the components of the \textit{SwiftTron} architecture, according to the operations needed. Its functionality is depicted in \Cref{fig:Control_Unit}. For the three major operations, which are MHSA, LayerNorm, and FFN, dedicated Finite State Machines (FSMs) generate the respective control signals. A set of handshake signals (e.g., $Start$, $Done$, $Valid$) is devised to interact between different FSMs and guarantee the correct timing of the operations in all stages of the Transformers' inference.

\section{Evaluation of our SwiftTron Architecture}
\label{sec:Evaluation}

\subsection{Experimental Setup}

We implement the complete design of our \textit{SwiftTron} architecture in RTL (VHDL) and evaluate it for the RoBERTa architecture~\cite{Liu_2019_RoBERTa} on the GLUE benchmark~\cite{Wang_2019_GLUE} and for the DeiT~\cite{Touvron_2021_DeiT} on the ImageNet dataset~\cite{Krizhevsky_2012_ImageNet}. We synthesize the \textit{SwiftTron} architecture in a $65\ nm$ CMOS technology node using the ASIC design flow with the Synopsys Design Compiler. We conduct functional and timing validation through gate-level simulations using Mentor Graphics QuestaSim. With the synthesized netlist, we obtain area, power, and performance of our design. We also run the inference on an Nvidia GeForce RTX 2080 Ti GPU for latency comparison.

\begin{figure}[t]
\centering
\includegraphics[width=.9\linewidth]{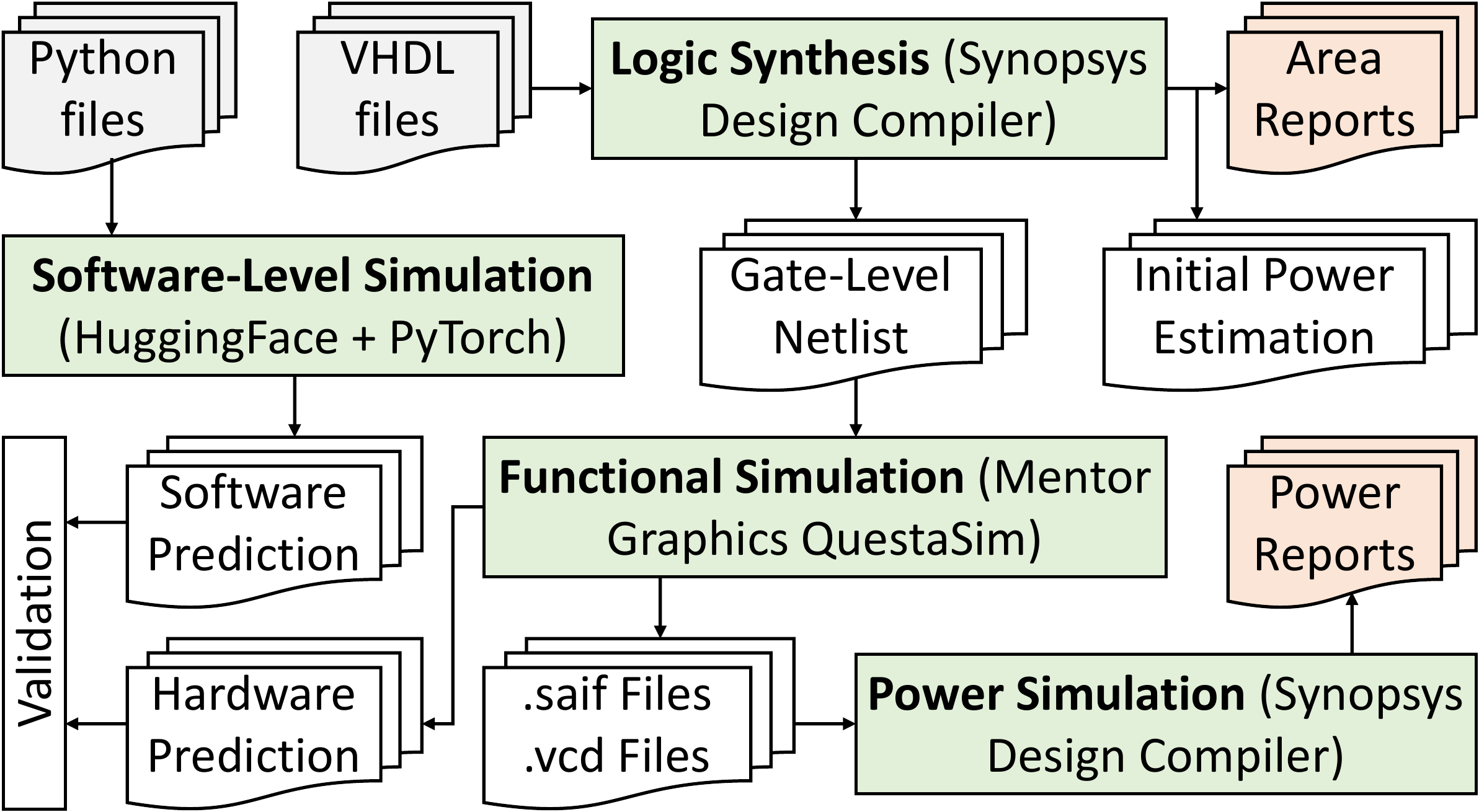}
\caption{Experimental setup and tool flow for conducting the experiments.}
\label{fig:Exp_setup}
\end{figure}

For validation, we use the pre-trained Transformer models from the HuggingFace library~\cite{Wolf_2020_HuggingFace}, and implement them on the PyTorch framework using the quantization algorithms of I-BERT~\cite{Kim2021iBert}. The complete flow is shown in \Cref{fig:Exp_setup}, where the grey boxes are the inputs, the orange boxes are the outputs, and the green boxes represent the main tools used. Note that this hardware design and validation flow are well-adopted by the hardware design community.

\subsection{SwiftTron Synthesis Results}

For evaluating the complete \textit{SwiftTron} design, we evluate the architecture with $d=768$, $k=12$, $m=256$, and $d_{ff}=3072$, to efficiently execute the RoBERTa-base architecture~\cite{Liu_2019_RoBERTa}. 
Note that these values are arbitrary parameters in our design and can be tuned during design time to support the execution of different Transformer architectures. The clock period has been set to $7\ ns$, which corresponds to a clock frequency of $\approx 143\ MHz$. To comply with the timing requirements, the computations of the Softmax and LayerNorm units have been partitioned into three pipeline stages.
Each component of the \textit{SwiftTron} architecture has been tested separately and simulated to validate its outputs compared to the software-level implementation of~\cite{Kim2021iBert}. Area and power consumption have been evaluated based on the reports obtained by the Synopsys Design Compiler tool, while the latency has been measured with a cycle-accurate simulator\footnote{The simulator considers the worst-case scenario in terms of clock cycles computed by the square root operator of the LayerNorm unit.}.
\Cref{tab:SwiftTron_Synthesis_Results} summarizes the key results obtained from the synthesis.  \Cref{tab:SwiftTron_speedup} reports the accuracy and latency of different Transformer models executed on our \textit{SwiftTron} architecture and the speedup of our accelerator compared to an Nvidia GeForce RTX 2080 Ti GPU with CUDA~$10$. Our experiments run more than $3 \times$ faster than the GPU implementations.

\begin{table}[t]
\centering
\caption{Summary of synthesis results of our proposed \textit{SwiftTron} architecture.}
\label{tab:SwiftTron_Synthesis_Results}
\resizebox{\columnwidth}{!}{%
\begin{tabular}{|c|c|c|c|c|}
\cmidrule{1-2} \cmidrule{4-5}
\textbf{Clock Frequency} & $143\ MHz$ & \hspace{5pt} & \textbf{Technology Node} & $65\ nm$ \\ \cmidrule{1-2} \cmidrule{4-5}
\textbf{Power Consumption} & $33.64\ W$ & \hspace{10pt} & \textbf{Area} & $273.0\ mm^2$ \\ \cmidrule{1-2} \cmidrule{4-5}
\end{tabular}%
}
\end{table}

\begin{table}[t]
\centering
\caption{Accuracy and inference latency for RoBERTa-base and RoBERTa-large models on the STT-2 task of the GLUE benchmark with sequence length $m=256$ and for DeiT-S on the ImageNet dataset with resolution $224 \times 224$ executed on our \textit{SwiftTron} architecture. The last column reports the speedup w.r.t. their execution on the RTX 2080 Ti GPU.}
\label{tab:SwiftTron_speedup}
\resizebox{\columnwidth}{!}{%
\begin{tabular}{c|c|c|c}
\textbf{Model} & \textbf{Accuracy} & \textbf{Latency} & \textbf{Speedup w.r.t. GPU} \\ \toprule
RoBERTa-base on STT-2 & $95.2\%$ & $1.83\ ms$ & $3.81 \times$ \\ \midrule
RoBERTa-large on STT-2 & $96.4\%$ & $45.70\ ms$ & $3.90 \times$ \\ \midrule
DeiT-S on ImageNet & $79.11\%$ & $1.13\ ms$ & $3.58 \times$
\end{tabular}%
}
\end{table}

\begin{table*}[t]
\centering
\caption{Summary of comparisons between the related works and our proposed \textit{SwiftTron} architecture.}
\label{tab:Related_Work_Comparison}
\begin{tabular}{c|c|c|c|c}
\textbf{Work} & \textbf{HW Implementation} & \textbf{Bit-width} & \textbf{Complete Architecture} & \textbf{Nonlinear Function Computation} \\ \toprule
OPTIMUS~\cite{MLSYS2020_903ce922} & \cmark\ ASIC 28 nm & INT16 & \xmark\ No & \xmark\ N/A \\ \midrule
A\textsuperscript{3}~\cite{Ham_2020_A3} & \cmark\ ASIC 40 nm & \cmark\ INT8 & \xmark\ No & \cmark\ Integers (approximated) \\ \midrule
FTRANS~\cite{Li_2020_FTRANS} & \cmark\ Xilinx FPGA & INT16 & \cmark\ Yes & \xmark\ Integers using FFT \\ \midrule
Lu et al.~\cite{lu2020hardware} & \cmark\ Xilinx FPGA & \cmark\ INT8 & \xmark\ No & \cmark\ Integers (approximated) \\ \midrule
EFA-Trans~\cite{Xin_2022_EFATrans} & \cmark\ Xilinx FPGA & \cmark\ INT8 & \cmark\ Yes & \xmark\ LUT \\ \midrule
FQ-BERT~\cite{Liu_2021_FQBERT} & \cmark\ Xilinx FPGA & \cmark\ INT8 & \cmark\ Yes & \xmark\ LUT \\ \midrule
Lin et al.~\cite{lin2020towards} & \xmark\ TITAN V GPU & \cmark\ INT8 & \cmark\ Yes & \xmark\ FP32 \\ \midrule
I-BERT~\cite{Kim2021iBert} & \xmark\ Tesla T4 GPU & \cmark\ INT8 & \cmark\ Yes & \cmark\ Integers (approximated) \\ \midrule
I-ViT~\cite{Li2022iViT} & \xmark\ RTX 2080 Ti GPU & \cmark\ INT8 & \cmark\ Yes & \cmark\ Integers (approximated) \\ \midrule
Transformer Engine~\cite{Nvidia_H100} & \cmark\ ASIC 4 nm (inside H100 GPU) & \xmark\ FP8 & \cmark\ Yes & \xmark\ FP16 / FP32 \\ \midrule
\textbf{SwiftTron (ours)} & \cmark\ \textbf{ASIC 65 nm} & \cmark\ \textbf{INT8} & \cmark\ \textbf{Yes} & \cmark\ \textbf{Integers (approximated)}
\end{tabular}%
\end{table*}

\subsection{Area and Power Breakdown}

\begin{figure}[t]
\centering
\includegraphics[width=.8\linewidth]{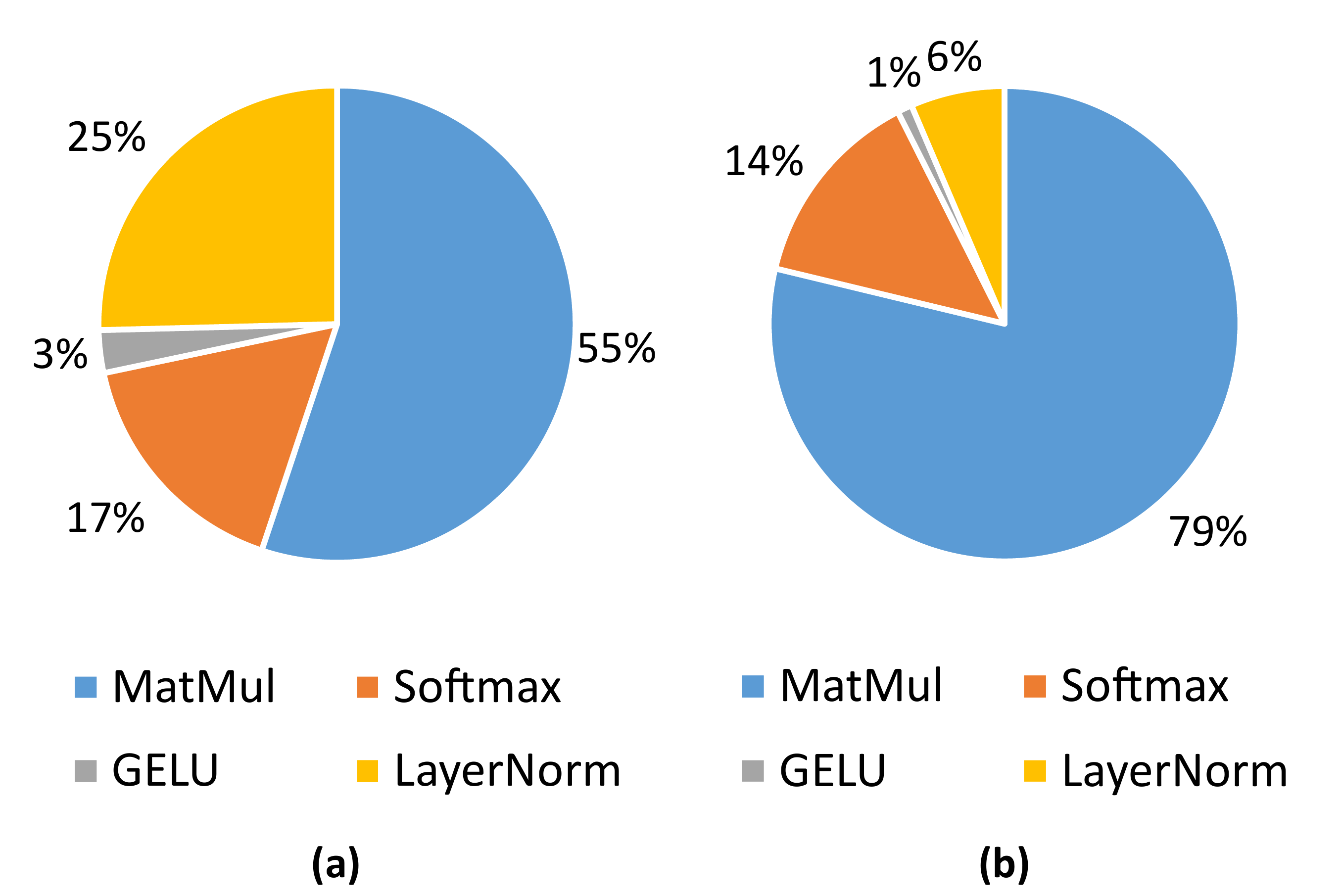}
\caption{\textbf{(a)} Area and \textbf{(b)} power breakdown of our \textit{SwiftTron} architecture.}
\label{fig:Area_Power_Breakdown}
\end{figure}

The total values of the area and power consumption of the complete \textit{SwiftTron} architecture are reported in the respective lines of \Cref{tab:SwiftTron_Synthesis_Results}, while \Cref{fig:Area_Power_Breakdown} analyzes in detail the breakdown for each component. It is evident that the MatMul block is responsible for the majority ($55\%$) of the area of the entire architecture. The difference between MatMul and other components becomes even larger for the power consumption. While the Softmax unit occupies $17\%$ of the total area, its contribution to the total power is only $14\%$. An even more significant difference is noted for the LayerNorm unit, which occupies $25\%$ area, but its power consumption is $6\%$. As expected, the GELU unit is a small component with only $3\%$ area and $1\%$ power consumption.

\subsection{Comparison with Related Works}
\label{subsec:Comparison_Related_Work}

While it is known that a specialized accelerator brings immense advantages compared to executing the same task on a general-purpose hardware device, e.g., CPU or GPU, it is difficult to compare metrics like power and performance across completely different hardware platforms of the related works, which include GPUs, FPGAs, and other ASIC architectures synthesized for different technology nodes. Hence, we identify key features of a hardware architecture for Transformers that push its efficiency to the upper boundary. Not only the hardware device on which it is implemented is important, but also the bit-width plays a key role in determining its energy efficiency. This is because even simple operators like adders and multipliers are relatively lightweight when implemented using integer arithmetic and low bit-width, like for INT8. On the other hand, their floating-point implementation incurs a significant complexity overhead. 

\Cref{tab:Related_Work_Comparison} summarizes the comparison between our \textit{SwiftTron} architecture and the related works, considering these important features for a hardware architecture for Transformers. From the table, it is clear that \textit{our work complies with all the requirements, while all the related works have at least one missing feature}. Some works~\cite{lin2020towards}\cite{Kim2021iBert}\cite{Li2022iViT} implement their design on GPUs, other works~\cite{MLSYS2020_903ce922}\cite{Ham_2020_A3}\cite{lu2020hardware} accelerate only part of a complete Transformer, and other works do not use efficient computations for their nonlinear functions. The design proposed in~\cite{Li_2020_FTRANS} uses integer computations, but its complexity is high due to the presence of FFT transforms. The architectures presented in~\cite{Xin_2022_EFATrans}\cite{Liu_2021_FQBERT} use LUTs for computing some nonlinear operations, while the works in~\cite{lin2020towards}\cite{Nvidia_H100} performs the nonlinear computations using FP16 or FP32 arithmetics.

\section{Conclusion}

In this paper, we present \textit{SwiftTron}, a specialized accelerator for Transformers that executes all the operations, including the nonlinear operations, in integer arithmetic. The correct computations between integers are achieved through a specialized quantization scheme that accounts for diverse scaling factors. Dedicated designs implement approximated versions of the nonlinear units, like Softmax, GELU, and Layer Normalization. The \textit{SwiftTron} architecture synthesized using the ASIC design flow shows efficient area, power, and performance while complying with all the desired features for a Transformer accelerator. Our design and thorough analyses pave the way for future developments of efficient Transformer architectures.


\section*{Acknowledgment}

This work has been supported in part by the Doctoral College Resilient Embedded Systems, which is run jointly by the TU Wien’s Faculty of Informatics and the UAS Technikum Wien.
This work was also supported in parts by the NYUAD’s Research Enhancement Fund (REF) Award on “eDLAuto: An Automated Framework for Energy-Efficient Embedded Deep Learning in Autonomous Systems”, and by the NYUAD Center for Artificial Intelligence and Robotics (CAIR), funded by Tamkeen under the NYUAD Research Institute Award CG010.

\bibliographystyle{ieeetr}
\bibliography{main.bib}

\end{document}